\documentclass[runningheads]{llncs}

\usepackage{eccv}

\usepackage{eccvabbrv}
\usepackage{multirow}
\usepackage{multicol}
\usepackage{graphicx}
\usepackage{booktabs}
\usepackage{algorithm}
\usepackage{algpseudocode}
\usepackage{arydshln}
\usepackage{float}

\usepackage[accsupp]{axessibility}  % Improves PDF readability for those with disabilities.

\usepackage{hyperref}

\usepackage{orcidlink}

\usepackage{xcolor}

\begin{document}

\title{Memory-Optimized Once-For-All Network}

\titlerunning{Memory-Optimized Once-For-All Network}

\author{Maxime Girard \and Victor Qu\'etu\orcidID{0009-0004-2795-3749} \and  Samuel Tardieu\orcidID{0000-0002-6885-1480} \and Van-Tam Nguyen \and Enzo Tartaglione\orcidID{0000-0003-4274-8298}}

\authorrunning{M.~Girard et al.}

\institute{LTCI, T\'el\'ecom Paris, Institut Polytechnique de Paris, France \email{\{name.surname\}@telecom-paris.fr}
}

\maketitle

\begin{abstract}
Deploying Deep Neural Networks (DNNs) on different hardware platforms is challenging due to varying resource constraints. Besides handcrafted approaches aiming at making deep models hardware-friendly, Neural Architectures Search is rising as a toolbox to craft more efficient DNNs without sacrificing performance. Among these, the Once-For-All (OFA) approach offers a solution by allowing the sampling of well-performing sub-networks from a single supernet- this leads to evident advantages in terms of computation. However, OFA does not fully utilize the potential memory capacity of the target device, focusing instead on limiting maximum memory usage per layer. This leaves room for an unexploited potential in terms of model generalizability.

In this paper, we introduce a \textbf{M}emory-\textbf{O}ptimized \textbf{OFA} (\textbf{MOOFA}) supernet, designed to enhance DNN deployment on resource-limited devices by maximizing memory usage (and for instance, features diversity) across different configurations. Tested on ImageNet, our MOOFA supernet demonstrates improvements in memory exploitation and model accuracy compared to the original OFA supernet. Our code is available at \url{https://github.com/MaximeGirard/memory-optimized-once-for-all}.\footnote{This work has been accepted for publication at the The Fourth Workshop on Computational Aspects of Deep Learning (CADL), and will be part of the ECCV 2024 Workshops proceedings.}

\end{abstract}

\section{Introduction}

Deep neural networks (DNNs) offer state-of-the-art performance in a range of machine learning applications. DNNs are being used in almost every industry, including video games~\cite{schiffer2023actor,defranco2023computing}, autonomous vehicles~\cite{kebria2019deep}, medical applications~\cite{abdelrahman2021convolutional}, chatbots such as OpenAI's ChatGPT, etc. These networks outperform the majority of other machine learning models on difficult tasks~\cite{issitt2022classification}: they are robust, highly effective, and generalizable~\cite{Guan_2020, rolnick2018deep} motivating their widespread use.

\begin{figure}[t]
    \centering
    \includegraphics[width=0.9\linewidth]{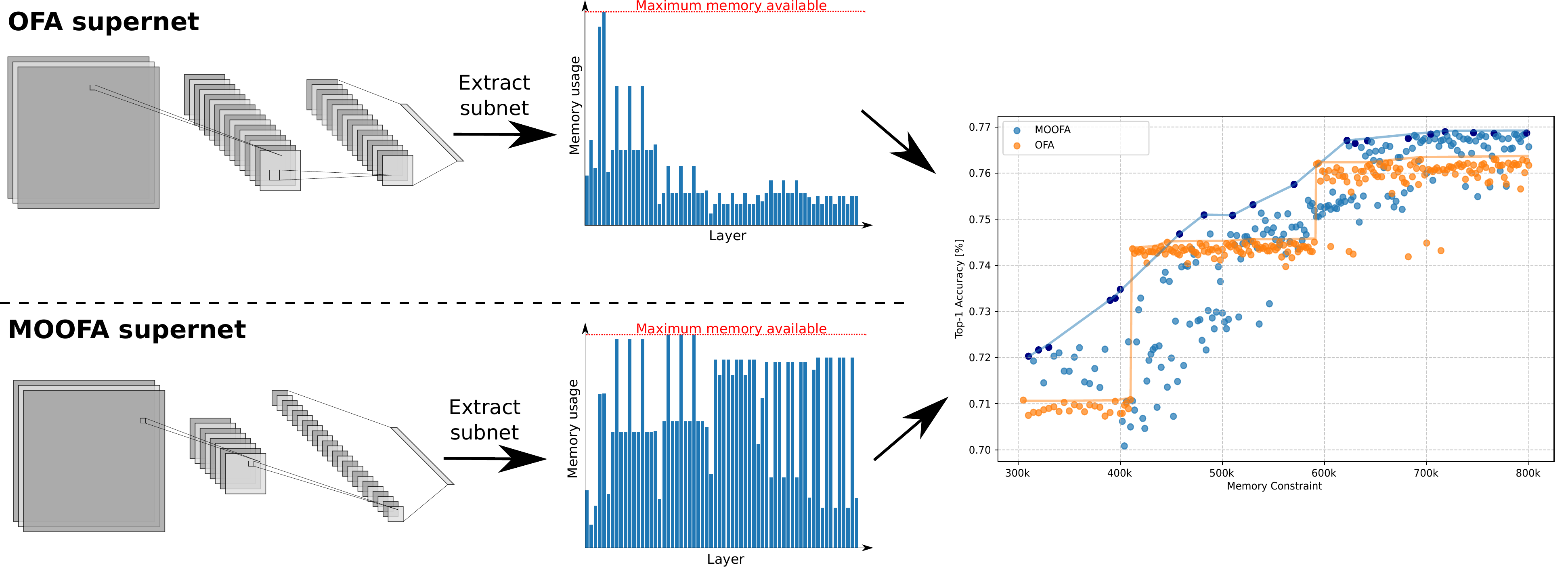}
    \caption{We propose a revised architecture of the OFA supernet aiming at providing memory-optimized subnets with improved accuracy under tight memory constraints.}
    \label{fig:visual_abstract}
\end{figure}

However, the size of these models has grown exponentially through the years~\cite{steiner2022olla}, leading to challenges in terms of memory and processing power needs~\cite{bragagnolo2022update}. One major obstacle to using DNNs on resource-constrained systems is the significant and ongoing increase in resource requirements. Moreover, deploying DNNs on diverse hardware platforms is a well-known issue. Indeed, it is necessary to optimize the model for each deployment scenario since various devices —from expensive smartphones to low-cost microcontrollers— have varying capacities and impose various constraints. However, identifying the best architecture for every hardware platform is very costly, both in terms of time and energy~\cite{strubell2020energy}.
It is primarily expensive because a specific solution needs to be designed for each deployment scenario, requiring more energy expenditures translated in CO$_2$ emissions~\cite{strubell2020energy}.
As the quantity and variety of IoT and embedded devices increase~\cite{ornes2016internet}, it is becoming important to come up with creative ways to create architectures that are effectively optimized for every deployment scenario.

The Once-For-All (OFA) network~\cite{cai2019once}, has been specially designed to meet this challenge. This method enables the selection of sub-networks designed for particular hardware constraints. Indeed, the OFA network is a once-trained supernet from which dynamic sub-networks are adapted for varying depth, width, kernel size, and resolution. The OFA network can effectively accommodate the unique limitations of different devices thanks to its adaptability and flexibility. The OFA network is trained to support a variety of sub-networks, offering that each sub-network operates best through a \emph{progressive shrinking} approach.

However, subnets from the OFA network have flaws. Specifically, models meeting memory requirements show a memory peak in the initial layers that significantly restricts the OFA search space. As shown in Fig.~\ref{fig:visual_abstract}, the layers after these initial ones typically use much less memory, leading to a suboptimal architecture in terms of generalizability. Allocating more memory in later layers would allow the extraction of more diverse features, boosting model performance.

In this paper, we propose a memory optimization management scheme that, given a memory consumption target, maximizes the per-layer features extracted, enabling a boost in the DNN performance. Through a decomposition of the memory contribution for each layer, we can size each layer's dimension while at the same time remaining under the target memory available. Unlike other OFA-based approaches that optimize the searched sub-networks in terms of performance and computation simply under a memory budget, MOOFA offers a perspective exploiting an unexpressed potential for the target network.

Here below we summarize our key messages and contributions.
\begin{itemize}
    \item We identify an inefficiency derived from the OFA supernet, where a few high-memory demanding layers constrain the selection for all the rest of the architecture. Precisely, layers close to the input are considerably more memory-demanding than deeper ones. This leads to an underfitting in terms of feature diversity (Sec.~\ref{subsec:OFA}).
    \item We decompose the memory consumption in every layer for the supernet, identifying recurring blocks in it. Through this analysis, we can properly size the layers accounting for the memory consumption of each (Sec.~\ref{subsec:channel}).
    \item When testing on ImageNet-1k (and ablating on one of its subsets, ImageNette), we observe that, especially in more memory-constrained scenarios, while remaining with the same memory constraint, we can better optimize the memory consumption (more features are extracted) which leads to a performance improvement (Sec.~\ref{sec:experiments}).
    
\end{itemize}
\section{Related Works}
\label{sec:sota}
\subsection{Neural Architecture Search}
\label{subsec:nas}
Neural Architecture Search (NAS) has become an important area for DNN architecture design. It seeks to automatically identify efficient architectures for tasks for which human expertise was previously required.
In a standard setup, NAS generally begins with a set of predefined operation sets and uses a controller to obtain a large number of candidate neural architectures based on the search space created by these operation sets. The candidate architectures are then trained on the training set and ranked according to their accuracy on the validation set. The ranking information of the candidate neural architecture is then used as feedback information to adjust the search strategy, enabling a set of new candidate neural architectures to be obtained. When the stop condition is reached, the search process is terminated. The chosen neural architecture then conducts performance evaluation on the test set. \\
Several methods have been developed to identify architectures. \cite{zoph2016neural, zoph2018learning} first used reinforcement learning to achieve this goal. LEMONADE~\cite{elsken2018efficient} is an evolutionary algorithm implementing Lamarckism: after every generation, child networks are generated to improve the Pareto-frontier concerning the current population. Other evolutionary algorithms use concepts like Montecarlo optimization~\cite{wang2021sample} or random search~\cite{li2020random}, which however significantly make the research of optimized architectures extremely difficult and complex to achieve, requiring thousands of computational days to optimize even on smaller datasets. The main obstacles to deploying these approaches are discussed below.\\
\noindent \textbf{The search space size.} This requires the NAS to use a search strategy that searches all necessary components of the neural architecture. This means that NAS needs to find an optimal neural architecture within a very large search space. The larger the search space, the higher the corresponding search cost.\\
\noindent \textbf{The search strategy is in general non-differentiable, or a proxy strategy}. This regards the differences between different neural architectures as a limited set of basic operations; that is, by discretely modifying an operation to change the neural architecture. This means that we cannot use the gradient strategy to quickly adjust the neural architecture.\\
DARTS \cite{liu2018darts} has been a ground-breaking proposal as it continuously relaxes the originally discrete search strategy, making it possible to use gradients to efficiently optimize the architecture search space. DARTS follows the cell-based search space of NASNet \cite{zoph2018learning} and further normalizes it by adding a softmax re-weighting which aids in the selection of the nodes to be included in the model.\\ 
\noindent \textbf{Full training is required for each candidate architecture.} In vanilla NAS, training each candidate neural architecture from scratch until convergence is typically required. This approach is sub-optimal: the network structures of the subsequent networks and previous ones are similar; therefore, it is clear that this relationship will not be fully utilized if each candidate neural architecture is trained from scratch. Also, we only need to obtain the relative performance ranking of the candidate architecture, and not necessarily the absolute performance for each proposed DNN. Whether it is necessary to train each candidate architecture to convergence is also a question worth considering.\\
Indeed, since NAS must be performed for every new hardware, it results in significant computational demands, as well as high costs and CO2 emissions~\cite{strubell2020energy}. 

\subsection{Once-For-All}
\label{subsec:OFA}

\begin{figure}[t]
\centering
\begin{subfigure}[b]{0.6\linewidth}
    \centering
    \includegraphics[width=\linewidth]{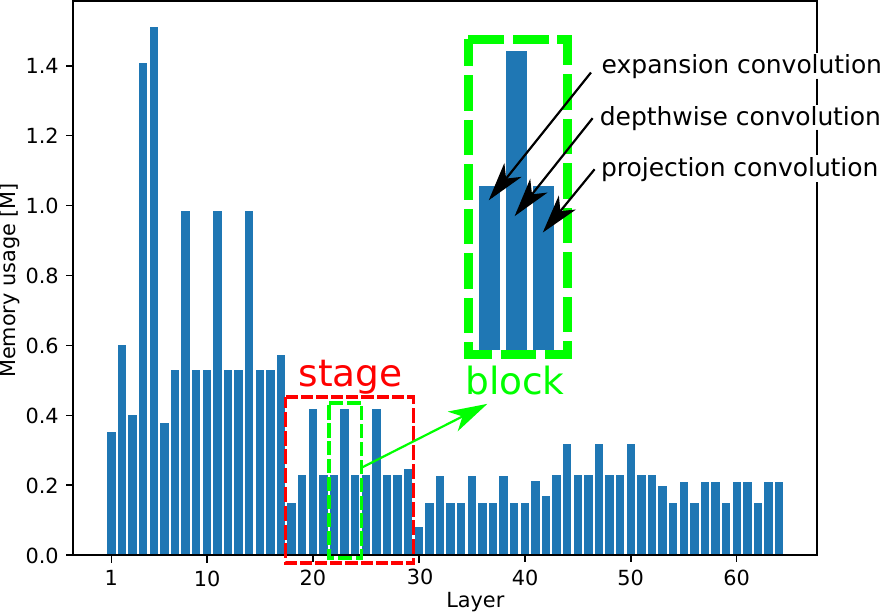}
    \caption{FLOPs and memory trends for ResNet-18.}
    \label{fig:memory_usage}
\end{subfigure}
\hfill
\begin{subfigure}[b]{0.35\linewidth}
    \centering
    \includegraphics[width=\linewidth]{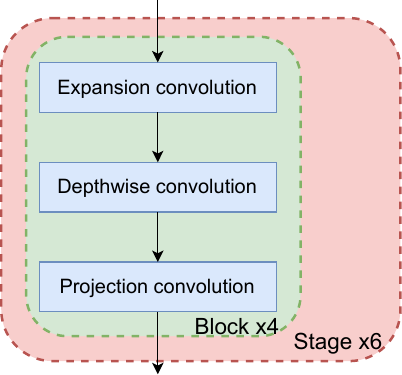}
    \caption{Structure of the OFA supernet (excluding the first and the last block).}
    \label{fig:structure}
\end{subfigure}
\caption{ (a) Memory usage of the original OFA network during a forward pass. and (b) internal structure of the OFA supernet. A memory peak resulting from the initial blocks significantly restricts submodels produced by OFA. The blocks that come after these initial ones therefore use much less memory.}
\label{fig:flops-memory-comparison}
\end{figure}

To overcome the latter, \cite{cai2019once} proposed to decouple the training and the search steps. To do so, they trained once a \textit{supernet}, which is a large model family ($2 \cdot 10^{19}$ subnets) using weight sharing, with Progressive Shrinking (PS). From the OFA supernet, we can dynamically extract sub-networks adapted for varying depth, width, kernel size, and resolution. The OFA network can effectively accommodate the unique limitations of different devices thanks to its adaptability. The OFA network is trained to support a variety of sub-networks, offering that each sub-network operates optimally through Progressive Shrinking.\\ 
Although the OFA approach greatly improves deployment efficiency and reduces its impact on the environment, our study points out areas that still require improvement, especially in the memory occupation of the sub-networks during inference. Indeed, Fig.~\ref{fig:memory_usage} shows the memory usage of a network produced by the original OFA supernet during a forward pass. We can observe a memory peak resulting from the initial layers which significantly restricts the models produced by OFA. The layers that come after these initial ones therefore use much less memory.
Our new architecture MOOFA aims at solving this memory usage problem by balancing the memory usage across the network while achieving better performance. 
\begin{figure}[t]
    \centering
    \includegraphics[width=0.75\linewidth]{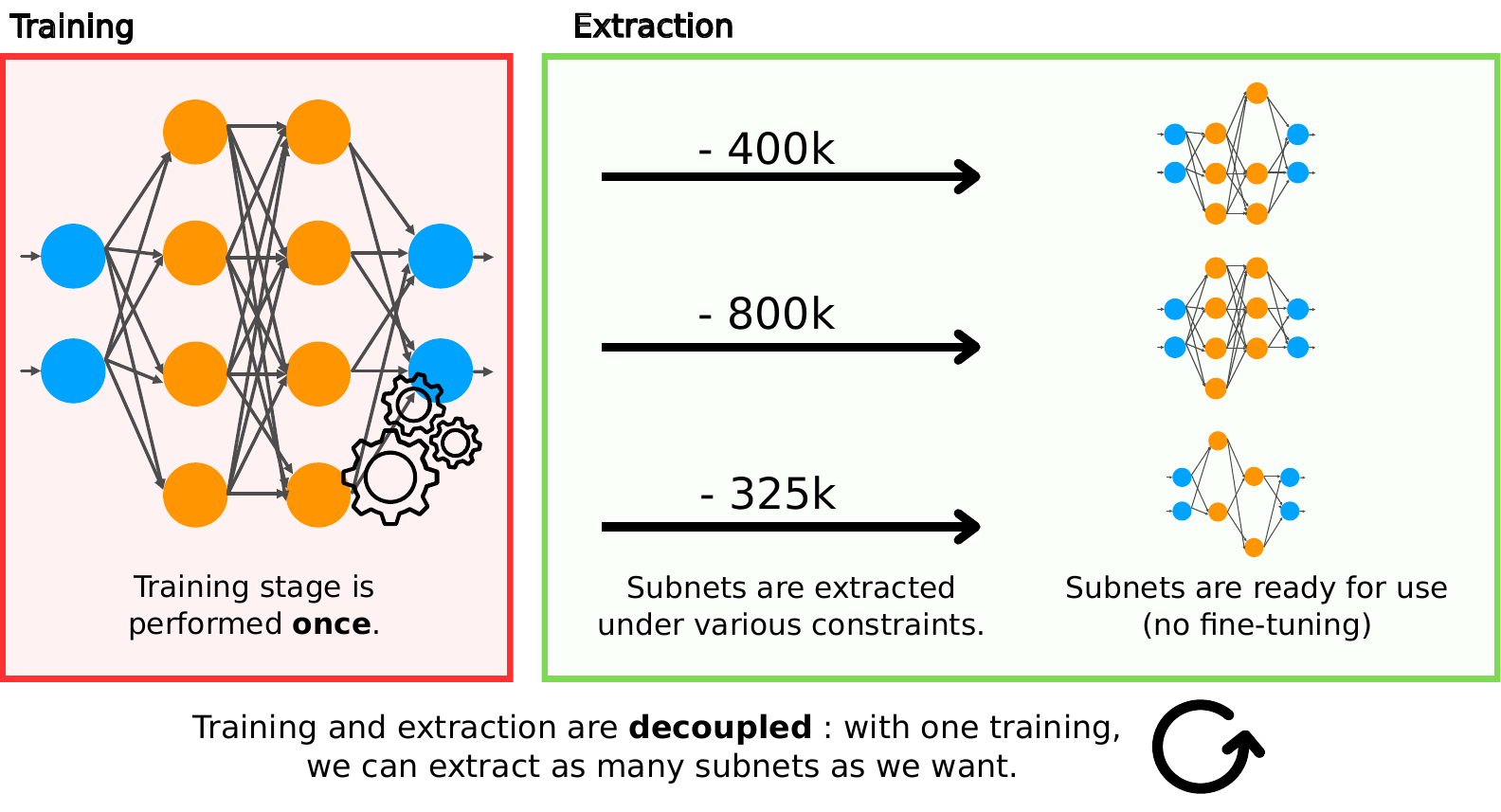}
    \caption{The Once-for-All production pipeline. A supernet is trained once, from which multiple subnets, at specific hardware constraints, can be drawn.}
    \label{fig:once-for-all}
\end{figure}
\cite{sahni2021compofa} proposed a new architecture space CompOFA, \emph{i.e.} a new supernet, to reduce its size by order of magnitude. They show that, in comparison to the OFA supernet, they can achieve a 216x speedup in model search and a 2x reduction in training time on ImageNet experiments, without sacrificing Pareto optimality. 
On top of this work, \cite{fangfast} go one step further and suggest an in-place knowledge distillation process to train the super-network and the sub-networks simultaneously. They create an upper-attentive sample technique that lowers the training cost per epoch without sacrificing accuracy within this in-place distillation framework. They show that in comparison to CompOFA, they can cut the training time by 1.5 times without sacrificing Pareto optimality. Furthermore, DetOFA~\cite{sakuma2023detofa} was designed as a method that uses search space pruning with relatively little training data, while TOFA~\cite{kundu2023transfer} utilized a unified semi-supervised training loss to simultaneously train all subnets within the supernet, coupled with on-the-fly architecture selection at deployment. While these methods present new sub-network selection strategies, our method does not aim at this but instead proposes a new architecture seeking optimal memory use.  

Built to improve the use of the memory, we will compare our new architecture MOOFA with the original OFA~\cite{cai2019once} and CompOFA~\cite{fangfast}.
\section{Method}
\label{sec:method}
Our main goal is to construct an OFA supernet with uniform memory usage throughout the network. To achieve this, critical points are taken into account, typically during stage transitions, when memory usage varies significantly. 
The key idea is to use as much memory as possible in every stage while staying below the previous stage's peak memory usage. With this approach, we can maximize the network's generalization capability while maintaining a constant overall memory requirement during inference.
Our approach starts by examining the memory usage patterns at each stage, taking into account the various layers' contributions (depthwise convolution, expansion, and projection). We determine the target channel size for the subsequent stage at each stage transition. This calculation estimates the expected peak in the next stage and attempts to match the peak memory usage of the current stage as closely as possible.

To target this, we first formalize the memory usage over a block of Mobile Inverted Bottlenecks (Sec.~\ref{subsec:problem}), and we establish formulas to calculate the target channel sizes, considering scenarios where memory usage may be dominated by different layers (Sec.~\ref{subsec:channel}).
Then, we define our MOOFA supernet architecture (Sec.~\ref{subsec:MOOFA}) and finally present our approach to search under constraint (Sec.~\ref{subsec:search}).

\subsection{Problem Formalization}
\label{subsec:problem}
First, we formalize the memory usage over a block of Mobile Inverted Bottlenecks (MB Blocks), which are the main components of the OFA supernet structure, as shown in Fig.~\ref{fig:structure}. Important parameters that are taken into account in our analysis are the stride $S$, the number of input channels $C_{in}$ and output channels $C_{out}$, the expand ratio factor $E$, the kernel convolution size $K$ and the input feature map size $I$. 
Three main components—the expansion convolution layer, the depthwise convolution layer, and the projection convolution layer—are examined in terms of memory usage during a forward pass.
For each layer in the MB block, the memory complexity is expressed here below.
The memory complexity of the expansion convolution layer is given by:
\begin{equation}
    M_{exp} = \underbrace{\mathcal{O}(C_{in}I^2)}_\text{input memory} + \underbrace{\mathcal{O}(C_{in}EC_{in})}_\text{weight memory} + \underbrace{\mathcal{O}(EC_{in}I^2)}_\text{output memory},
\end{equation}
where input memory stores the initial feature map, weight memory holds the learnable parameters and output memory stores the expanded feature map.
Further, the memory complexity of the depthwise convolution layer is defined as:
\begin{equation}
    M_{dw} = \underbrace{\mathcal{O}(EC_{in}I^2)}_\text{input memory} + \underbrace{\mathcal{O}(EC_{in}K^2)}_\text{weight memory} + \underbrace{\mathcal{O}\left[EC_{in}\left(\frac{I}{S}\right)^2\right]}_\text{output memory},
\end{equation}
where input memory stores the expanded feature map, weight memory contains the depthwise convolution kernels and output memory holds the spatially convolved features.
Furthermore, the projection convolution layer's memory complexity is expressed by
\begin{equation}
    M_{proj} = \underbrace{\mathcal{O}\left[EC_{in}\left(\frac{I}{S}\right)^2\right]}_\text{input memory} + \underbrace{\mathcal{O}(EC_{in}C_{out})}_\text{weight memory} + \underbrace{\mathcal{O}\left[C_{out}\left(\frac{I}{S}\right)^2\right]}_\text{output memory},
\end{equation}
where input memory stores the depthwise convolved features, weight memory holds the projection convolution parameters, and output memory contains the final output feature map. 
Note that within a stage (where $S=1$ and $C_{in}=C_{out}$), we observe $M_{exp} = M_{proj}$.

\subsection{Channel Size Optimization for Memory-Constant Architecture}
\label{subsec:channel}
Since our goal is to optimize channel sizes across stages for consistent peak memory usage, three scenarios to determine the memory peak are considered.

\noindent\textbf{Depthwise-Dominated Scenario.}
Assuming the depthwise convolution layer dominates memory usage in both the current and next stages, we can write:
\begin{equation}
    M_{dw}(C_{in}) = M_{dw}(C_{out}).
\end{equation}
By expanding and solving for $C_{out}$, it results
\begin{equation}
    C_{out} = C_{in} \cdot \frac{I^2 + 4K^2 + \frac{I^2}{4}}{2I^2 + 4K^2}.
\end{equation}

\noindent\textbf{Depthwise to Expansion Transition.}
If the peak arises during transitions from depthwise to the expansion layer, we have
\begin{equation}
    M_{dw}(C_{in}) = M_{exp}(C_{out}),
\end{equation}
leading to a quadratic equation in $C_{out}$, which gives us the solution of $C_{out}$ as
\begin{equation}
    C_{out} = \frac{-(\frac{EI^2}{4} + \frac{I^2}{4}) + \sqrt{(\frac{EI^2}{4} + \frac{I^2}{4})^2 - 4E[-EC_{in}(I^2 + K^2 + \frac{I^2}{4})]}}{2E}.
    \label{eq:Cout_depthwise_to_expansion_transition}
\end{equation}

\noindent\textbf{Expansion-Dominated Scenario.}
If the expansion layer dominates in the current stage, there is only one possibility for the next stage: it must be dominated by the depthwise convolution layer. Hence, we can write
\begin{equation}
    M_{exp}(C_{in}) = M_{exp}(C_{out}),
\end{equation}
which also results in a quadratic equation in $C_{out}$, therefore the solution follows the same form as in the previous scenario in \eqref{eq:Cout_depthwise_to_expansion_transition}. 

\noindent\textbf{Optimal Channel Size Selection.} Finally, to determine the optimal $C_{out}$, the appropriate value based on the current stage's dominant layer is selected:
\begin{equation}
    C_{out}^* = \begin{cases}
        \min(C_{out,dw}, C_{out,dw-exp}) & \text{if depthwise dominates current stage} \\
        C_{out,exp-dw} & \text{if expansion dominates current stage},
    \end{cases}
\end{equation}
where the indices $dw$ and $dw-exp$ refer respectively to the depthwise-dominated scenario and the depthwise to expansion transition.
This approach ensures minimal memory usage while maintaining consistency across stages, taking into account the realistic transitions between dominant layers in consecutive stages.
Hence, we define a new OFA supernet architecture in the next subsection.

\subsection{Memory-Optimized OFA (MOOFA)}
\label{subsec:MOOFA}

To optimize memory usage, we propose a new OFA architecture based on the original OFA architecture, but incorporating several modifications.

First, the same number of stages (5) and layer per stage options (2 to 4), the kernel size possibilities $K\in\{3, 5, 7\}$, and the input image size options [128, 160, 192, 224] are kept identical.
Moreover, the expand ratio factor E options are modified from [3, 4, 6] to [2, 3, 4] to reduce overall memory consumption. Furthermore, we propose a systematic method to determine channel sizes for each stage to design a memory-constant model. 

%Indeed, a reference configuration with particular parameters for the expand ratio factor, kernel size, and input image size is established in order to implement this approach and is presented in Appendix~\ref{appendix:reference_configuration}. 
The previously derived equations and this reference configuration allow us to compute a range of channel sizes for every network stage. The channel sizes that have been calculated for the selected reference configurations are as follows: 8, 24, 96, 288, 360, 384, 392, 392. The size of this family of subnets is $[(3 \cdot 3)^2 + (3 \cdot 3)^3 + (3 \cdot 3)^4]^5 \approx 2 \cdot 10^{19}$ models.

Please note that the significant memory peak induced by the final expand layer is addressed by calculating a final channel size using the same equations as before. Although it may impact performance, this change is required to keep memory constancy across the network.

\subsection{Search Under Constraint}
\label{subsec:search}

We developed a memory occupation logger computing the memory occupied by each layer during a forward pass, and allowing us to find the memory peak of a given configuration. Our code is publicly available. A dataset of randomly drawn configurations and their evaluated accuracies is also used to train an accuracy predictor. However, we must ensure sample balance and equal representation of all memory peak dynamics when constructing the dataset for the accuracy predictor.
This is because configurations satisfying tight constraints are less common in proportion, which reduces the likelihood that they will be chosen by random sampling.
To allow efficient searching under tight constraints, our predictor must also work well for configurations with smaller memory peaks.
The ability to optimize for tight constraint searches would be hampered if we only used random samples.
The search process is greatly sped up by this predictor, which enables quick inference of expected accuracy for a given configuration. 

We can quickly find the best subnet configurations that maximize expected performance while satisfying specific memory constraints by combining these two tools: the accuracy predictor and the memory estimator.
We use an evolutionary search algorithm to perform the search.

\section{Experiments}
\label{sec:experiments}

\subsection{Experimental Setup}
To assess the effectiveness of our proposed architecture, experiments using the ImageNet dataset are carried out. We compare our results with the original OFA architecture~\cite{cai2019once} as well as the CompOFA architecture~\cite{sahni2021compofa}. Progressive Shrinking was used to train the new architecture. 8 NVIDIA GeForce RTX 3090 GPUs were used for training, and PyTorch 2.3.1 and Horovod 0.28.1 were used. The pre-trained sources from which the other models were taken are mentioned in the corresponding papers. Other supernets are also trained on ImageNet.

\subsection{Preliminary Experiment}

We justify here the expand ratio factor options selected in Section~\ref{subsec:MOOFA} in this section. By training a memory-constant OFA supernet with configurations having fixed kernel sizes to [3,5,7], fixed depths to [2,3,4], and fixed input image size options [128, 160, 192, 224], we carried out a comparative study to find the most suitable setup. Three different expand ratios were tested: [3,4,6], [1, 1.5, 2], and [2, 3, 4] on Imagenette~\cite{imagenette2019}, a subset of ImageNet with 10 classes, to allow faster, more economical, and environmentally friendly ablation studies.

\iffalse
\begin{table}[t]
\centering
\setlength{\tabcolsep}{6pt}
\begin{tabular*}{\textwidth}{@{\extracolsep{\fill}}lccccc}
\hline
\textbf{Configuration} & \textbf{Kernel sizes} & \textbf{Depths} & \textbf{Expand ratios} & \textbf{Image sizes} \\
\hline
Configuration 1 & [3, 5, 7] & [2, 3, 4] & [3, 4, 6] & [128, 160, 192, 224] \\
\hline
Configuration 2 & [3, 5, 7] & [2, 3, 4] & [1, 1.5, 2] & [128, 160, 192, 224] \\
\hline
Configuration 3 & [3, 5, 7] & [2, 3, 4] & [2, 3, 4] & [128, 160, 192, 224] \\
\hline
\end{tabular*}
\caption{Different configurations we tested.}
\label{tab:ablation_conf}
\end{table}
\fi 

\begin{figure}[t]
        \centering
        \includegraphics[width=0.71\linewidth]{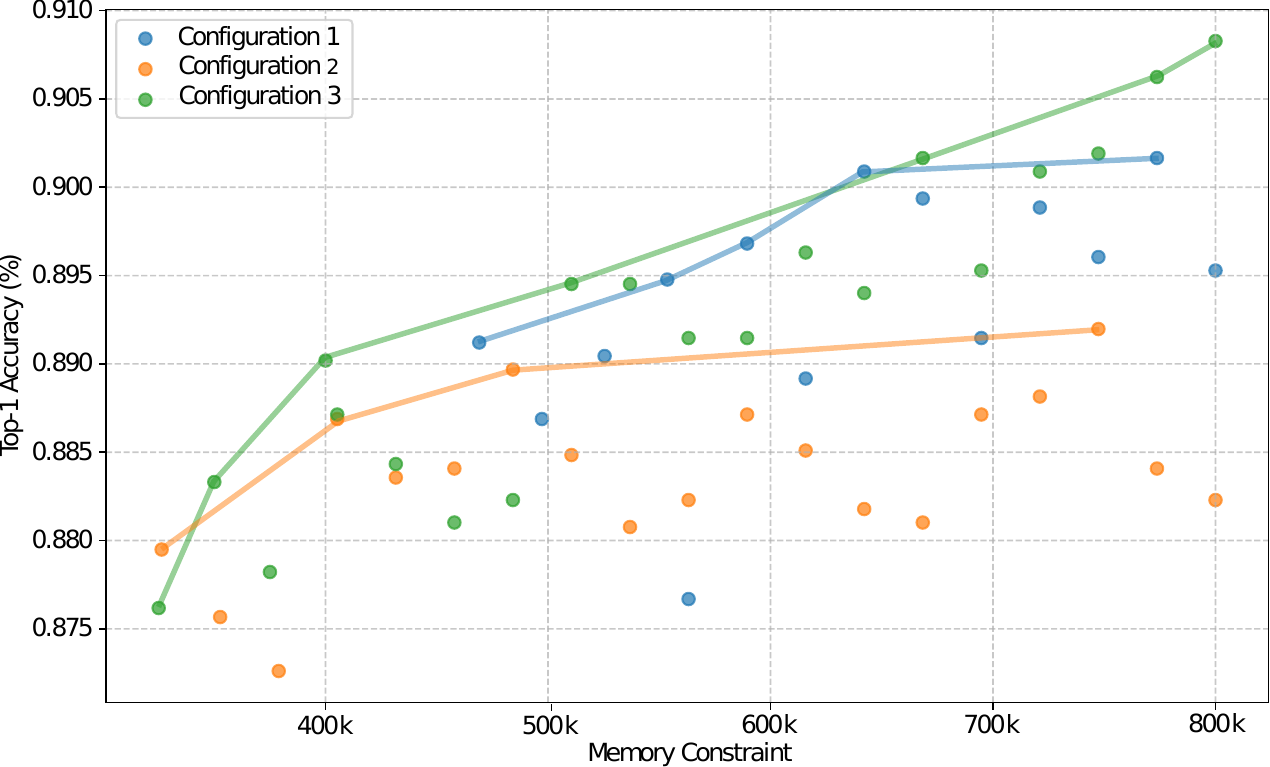}
    \caption{Top-1 accuracy under different memory constraints for three configurations trained on Imagenette. Configuration 1 uses the expansion ratio factor set from the original OFA ([3,4,5]). Configuration 2, respectively Configuration 3, use [1,1.5,2], respectively [2,3,4], as expansion ratio factor set.}
    \label{fig:ablation}
\end{figure}

The top-1 accuracy [\%] of found subneworks under specific constraints is presented in Fig.~\ref{fig:ablation}. The constraint is given in the maximum number of items in RAM at any given point during inference, as explained in Sec~\ref{exp:memory}. Configuration 1, which uses the expansion ratio factor set from the original OFA supernet ([3, 4, 6]), does not allow our new memory-constant architecture to find subnetworks with a memory peak usage lower than approximately 450k items in RAM.
It appears that the best configuration among the ones we explored was Configuration 3, using [2, 3, 4] as the expansion ratio factor. Indeed, this configuration allows both good performance at lower constraints (compared to Configuration 2), and the possibility for the supernet to produce models with a low memory peak.

\subsection{Memory Usage Exploration}
\label{exp:memory}
Through the use of a specially built logger that tracks each layer's memory usage during the forward pass, we were able to record memory usage.
The total amount of memory used by the inputs, outputs, and weights is included in the memory usage for each layer. This logger originally revealed information about the memory behavior of the first OFA supernet.
We effectively see how memory usage is distributed through a forward-pass, and how it could be optimized. 

We assume that the weights of the subsequent layers are sequentially loaded and unloaded from RAM, using a different storage medium (Flash, ROM), in order to optimize memory during inference. RAM is only used to store inputs and outputs for as long as is required. One image at a time is processed (batch size of 1). The chosen batch size is consistent with our objective of finding networks compatible with situations where memory is severely constrained. In such scenarios, the main obstacle to employing larger batch sizes (defined here as more than a few instances) is the inability to load the data into memory, as well as the results of subsequent computations, whose size increases linearly with batch size. Although lowering image resolution could allow for larger batch sizes, this strategy poses a significant risk of severely affecting accuracy, which is a critical concern we aim to avoid.

We tracked the memory usage for a forward pass through multiple subnets that were taken from our supernet under different constraints using the same logger, and we compared the results with subnets that were taken from the original OFA. The results confirm that there has been a great improvement in the distribution of memory usage.
\begin{figure*}[t]
    \centering
    \begin{subfigure}{0.49\textwidth}
        \centering
        \includegraphics[width=0.88\textwidth]{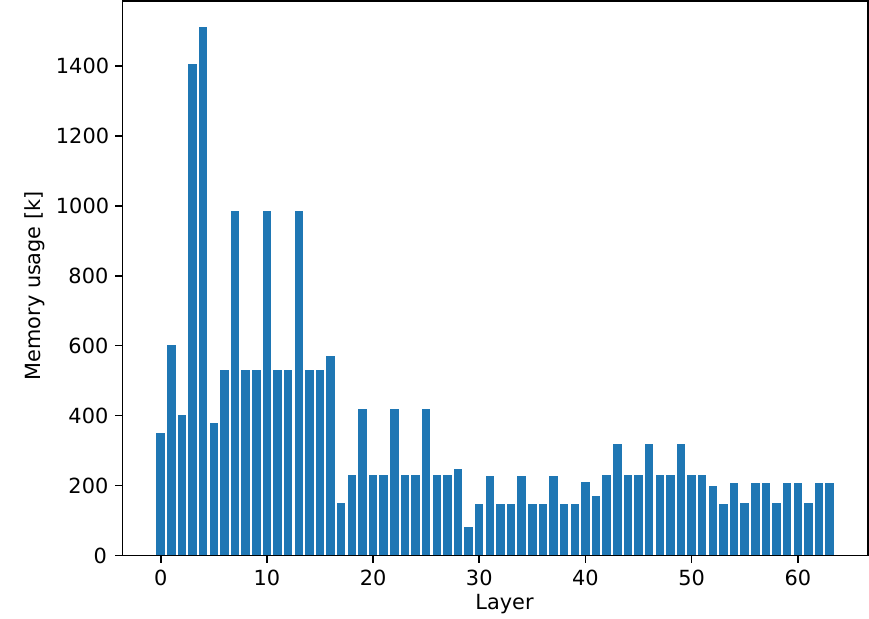}
        \subcaption{OFA - without constraint.}
    \end{subfigure}
    \begin{subfigure}{0.49\textwidth}
        \centering
        \includegraphics[width=0.88\textwidth]{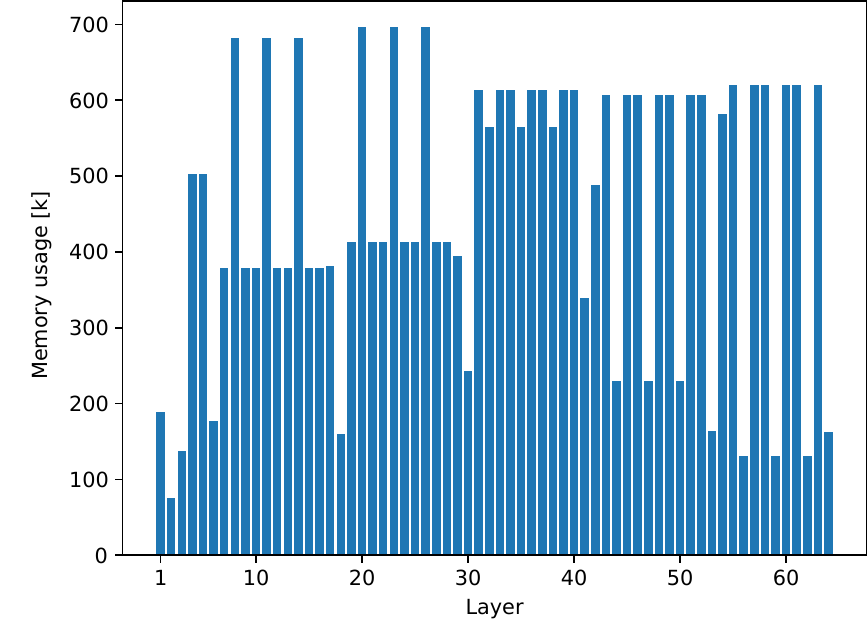}
        \subcaption{MOOFA (ours) - without constraint.}
    \end{subfigure}
    \\
    \begin{subfigure}{0.49\textwidth}
        \centering
        \includegraphics[width=0.88\textwidth]{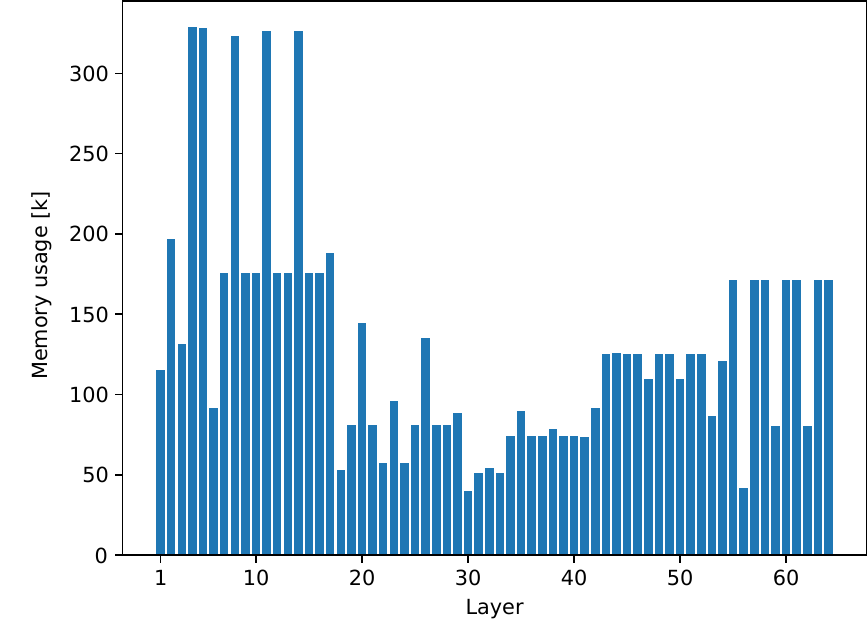}
        \subcaption{OFA - under 350k constraint.}
    \end{subfigure}
    \begin{subfigure}{0.49\textwidth}
        \centering
        \includegraphics[width=0.88\textwidth]{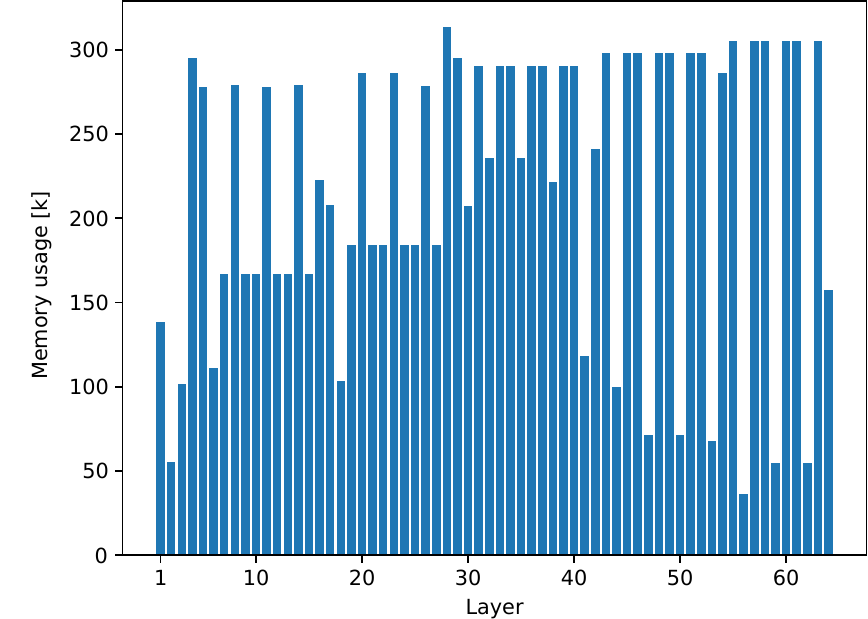}
        \subcaption{MOOFA (ours) - under 350k constraint.}
    \end{subfigure}
    \caption{Memory usage for a forward pass on OFA and MOOFA. With or without any constraint, while the original OFA architecture (a, c) presents a memory peak resulting from the initial blocks, our architecture MOOFA equalizes the memory usage across the entire network (b,d).}
    \label{fig:mem_results}    
\end{figure*}
Specifically, Fig.~\ref{fig:mem_results} depicts the memory usage for a forward pass for the original OFA, compared with our architecture MOOFA.
As previously discussed, whether under constraint or not, the original OFA displays a memory peak in the early stages, constraining the memory usage of subsequent layers (a,c).
However, the memory usage of MOOFA (b,d) is not strictly constant across the network but rather evenly distributed throughout the network, whether under constraint or not.

\subsection{Performance Comparison}
\begin{table}[t]
\centering
\begin{tabular*}{\textwidth}{@{\extracolsep{\fill}}lcccc}
\toprule
\textbf{Constraint} & \textbf{Metric} & \textbf{OFA} & \textbf{CompOFA} & \textbf{MOOFA(Ours)} \\
\midrule
 \multirow{4}{*}{Under 325k}& Top-1 Accuracy [\%] ($\boldsymbol{\uparrow}$) & 70.81 & 69.91 & \bf 72.13 \\
 & Memory peak [k] ($\boldsymbol{\uparrow}$) & 262.91 & 262.91 & \bf 305.28 \\
 & Avg. memory [k] ($\boldsymbol{\uparrow}$)& 119.68$\pm$51.52& 109.46$\pm$54.18 &\bf 214.92$\pm$84.30\\
 & FLOPs [M] ($\boldsymbol{\downarrow}$) & 146 & \bf 132 & 422 \\
\midrule
 \multirow{4}{*}{Under 350k}& Top-1 Accuracy [\%] ($\boldsymbol{\uparrow}$) & 70.91 & 70.26 & \bf 72.25 \\
 & Memory peak [k] ($\boldsymbol{\uparrow}$) & 328.70 & \bf 330.82 & 313.34 \\
 & Avg. memory [k] ($\boldsymbol{\uparrow}$) & 130.62$\pm$71.50& 115.79$\pm$65.77 &\bf 219.17$\pm$84.24\\
 & FLOPs [M] ($\boldsymbol{\downarrow}$) & 163 & \bf 134 & 431 \\
\midrule
 \multirow{4}{*}{Under 400k}& Top-1 Accuracy [\%] ($\boldsymbol{\uparrow}$) & 70.98 & 70.23 & \bf 73.34 \\
 & Memory peak [k] ($\boldsymbol{\uparrow}$) & 330.82 & 329.28 & \bf 381.12 \\
 & Avg. memory [k] ($\boldsymbol{\uparrow}$) & 130.62$\pm$72.07 & 123.18$\pm$74.75 & \bf 258.74$\pm$93.04\\
 & FLOPs [M] ($\boldsymbol{\downarrow}$) & 163 & \bf 144 & 554 \\
\midrule
 \multirow{4}{*}{Under 800k}& Top-1 Accuracy [\%] ($\boldsymbol{\uparrow}$) & 76.04 & 74.78 & \bf 76.58 \\
 & Memory peak [k] ($\boldsymbol{\uparrow}$) & \bf 740.42 & 722.45 & 686.98 \\
 & Avg. memory [k] ($\boldsymbol{\uparrow}$) & 212.08$\pm$155.57 & 200.11$\pm$136.77 & \bf 389.24$\pm$179.68\\
 & FLOPs [M] ($\boldsymbol{\downarrow}$) & \bf 283 & 306 & 959\\
\bottomrule
\\
\end{tabular*}
\caption{Top-1 Accuracy [\%], Memory peak, Average Memory, and FLOPs under different constraints for OFA, CompOFA, and MOOFA trained on ImageNet.}
\label{tab:results}
\end{table}

Table \ref{tab:results} presents the results in terms of top-1 accuracy [\%], memory peak, and FLOPs under different constraints for OFA, CompOFA, and MOOFA trained on ImageNet. We evaluated each OFA supernet's ability to provide effective sub-networks under various maximum memory constraints by conducting multiple searches.
The maximum memory usage of the inputs, outputs, and weights at any layer—expressed in terms of the number of items—defines the constraint. This implies that the total amount of memory used by the inputs, outputs, and weights of each layer must be less than this threshold. These values, expressed in the number of items, should be multiplied by 4 to convert them to the actual constraint in bytes because we utilized float32 precision for our experiments.

Please note that we omitted the memory usage for the last layer —the final expansion layer, a large linear layer that classifies among ImageNet's 1000 classes— from our studies since it made it more difficult to establish meaningful comparisons between the various OFA strategies. This layer was far bigger than any previous layer in the initial OFA setups, frequently turning into the "memory peak." This meant that the metric for peak memory usage was determined almost exclusively by this layer, making it hard to evaluate the impact of architectural optimizations on overall memory efficiency. In nearly all cases, this removal has no effect on our conclusions: the last layer is usually replaced for the specific classification task in real-world applications. For instance, real-world applications typically include much fewer classes than the 1,000 supported by the initial configuration (as in the whole ImageNet dataset), which can range from two in binary classification to a few dozen. In these scenarios, the last layer's memory requirements are low, accounting for only a few kilobytes of the model's total memory usage. However, the reader should be aware that if the full 1,000-class ImageNet setup is used (or any other task with hundreds of classes), the last layer may become the memory bottleneck due to its significantly higher memory requirements, which is in the hundreds of kilobyte range.

Whether under tight constraints (<400k) or large ones (<800k), our architecture MOOFA yields better performance in most cases, or similar in others, compared to the original OFA architecture and the CompOFA method. Indeed, under severe constraint, MOOFA showcases a performance gain of more than 2\% to CompOFA, and more than 1\% compared to the original OFA. 

However, everything comes at a cost, the number of floating points operations (FLOPs) exhibited by our architecture MOOFA is consistently higher than for the original OFA and CompOFA. Indeed, more features are extracted which leads to higher generalization but requires a lot more computation. In addition to FLOPs, we examined the relationship between execution time and supernet architecture. Findings indicate that execution time is more affected by depth than by parameters like channel sizes, likely due to GPU parallelism. As a result, even while our tests on a GPU reveal little time differences between the various supernet architectures, the same findings might be very different on an MCU because parallelism is less effective there.

Looking at the average memory, we can clearly see that MOOFA yields better results: since the memory is better distributed across all layers, the average memory is higher compared to the other approaches. Moreover, despite a better equalization across the full network has been achieved with MOOFA, we observe in most cases a lower memory peak compared to the other two approaches. Indeed, a better memory distribution across the network can lead to lower memory peaks as they should not constrain the memory usage of the following layers.

\iffalse
\begin{figure}[t]
    \centering
    \begin{minipage}{0.8\textwidth}
        \centering
        \includegraphics[width=\linewidth]{figures/accuracy_coparison.pdf}
    \end{minipage}
    \caption{Top-1 Accuracy achieved vs max number of parameters constraint for both OFA nets.}
    \label{fig:acc_constraints}
    \centering
    \begin{minipage}{0.8\textwidth}
        \centering
        \includegraphics[width=\linewidth]{figures/accuracy_comparison_tight_constraint.pdf}
    \end{minipage}
    \caption{Top-1 Accuracy achieved vs max number of parameters constraint for both OFA nets, focused on tighter constraints.}
    \label{fig:acc_constraints_tight}
\end{figure}
\fi
\begin{figure}[t]
    \centering
    \includegraphics[width=0.7\linewidth]{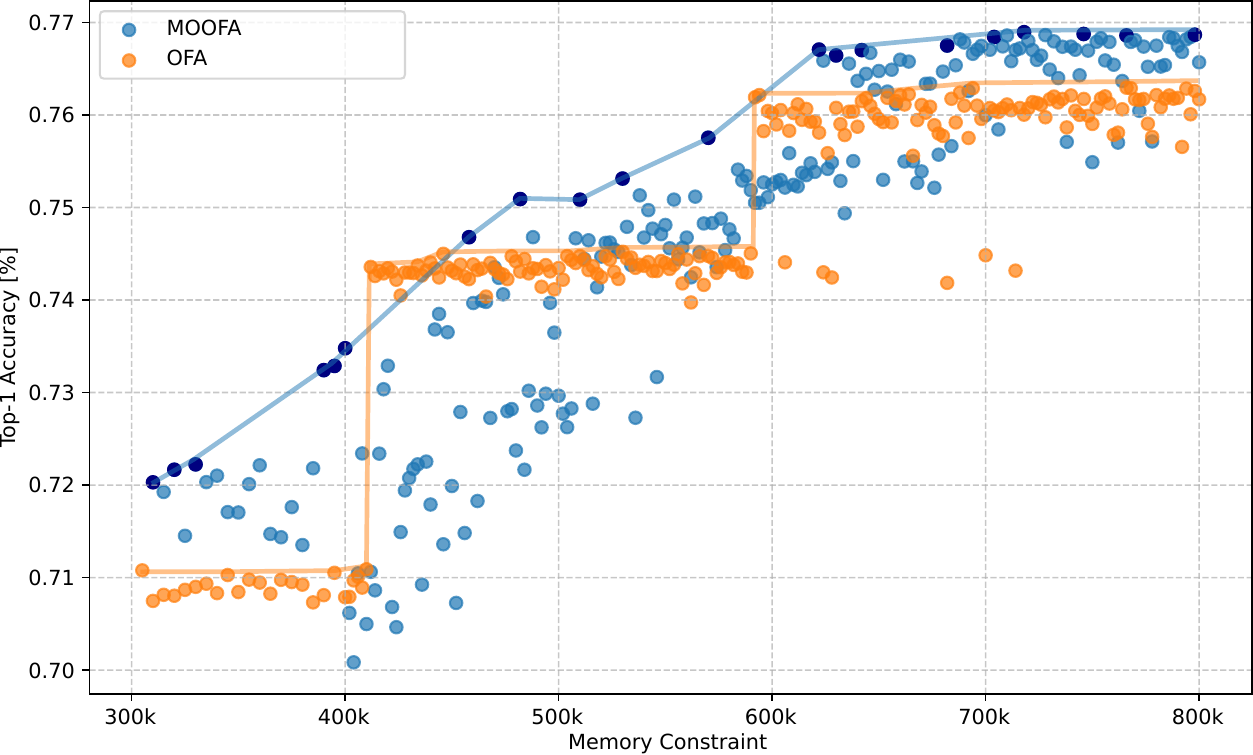}
    \caption{Top-1 accuracy achieved under different constraints for the original OFA supernet and our MOOFA architecture trained on ImageNet. The blue and orange segments delineate the best results attained by the two architectures.}
    \label{fig:acc_constraints}
\end{figure}
\iffalse
\begin{figure}[t]
    \centering
    \includegraphics[width=0.95\linewidth]{figures/accuracy_comparison_tight_constraint.pdf}
    \caption{Top-1 accuracy under severe memory constraint ([300k, 400k]) for the original OFA supernet and our MOOFA architecture trained on ImageNet.}
    \label{fig:acc_constraints_tight}
\end{figure}
\fi
Moreover, Fig.~\ref{fig:acc_constraints} displays the top-1 accuracy achieved by subnetworks taken from the supernetwork for the range of constraints we considered for both the original OFA and MOOFA.
Subnets extracted from MOOFA perform marginally better than the original OFA subnets under a permissive constraint (greater than 650k). Under severe constraints, looking at the interval [300k, 400k], the gap is larger. Indeed, MOOFA can find subnets that perform between 1 to 2\% better under the same memory constraint, hence achieving a Pareto frontier.
Our architecture MOOFA is achieving the Pareto Frontier in this setup.

\subsection{Discussion}
Our novel MOOFA architecture enables us to maintain good performance under severe constraints. In contrast to the original supernet, our approach targets memory-constant usage (in the ideal case), allocating more memory to the later stages. More resources improve our model's ability to generalize. Hence, our model can achieve similar accuracy under much stricter constraints by using the same cumulative amount of memory as a model generated by the original OFA supernet under larger constraints.

However, looking at the results within the 400k–500k memory constraint range, two things are apparent in this interval. First, the performance is noisy; second, we have identified subnets that achieve up to 1\% accuracy gain over OFA, along with some subnets that perform poorly. All things considered, our model enables reaching the Pareto frontier in the great majority of constraints.

Even though we aim to maximize memory occupation through the feed-forward process, we have to keep in mind that maintaining/boosting performance with stricter constraints comes at a cost. Indeed, for a given constraint, memory is globally more used.
The computing requirement is also affected: much more processing power is required to produce subnets with similar performance under stricter limitations.
When deploying models extracted by our MOOFA supernet, this must be taken into consideration as it affects both the latency of the sub-network but also the device's energy consumption. 
\section{Conclusion}

Based on the Once-For-All (OFA) supernet design, we have proposed in this work MOOFA, a novel memory-optimized supernetwork, designed to enhance DNN deployment on resource-constrained devices by maximizing memory usage across the network. The effectiveness of MOOFA has been demonstrated on ImageNet where our architecture showcases improvements in memory exploitation and model performance compared to the original OFA supernet. Our work shows that effective memory management can improve neural network performance in resource-constrained environments, but everything comes at a cost and these networks with optimized memory have higher computational overhead.

Future work will enhance MOOFA in two aspects. First, there is still an opportunity to optimize memory usage across layers for diverse subnet configurations. This requires defining a metric (e.g., entropy-based) to quantify memory usage consistency and developing an algorithm to explore the OFA architecture space. Second, co-training the predictor alongside the OFA network during progressive shrinking will reduce data imbalance by focusing on configurations considered during training, leading to a more efficient and adaptive process.

Concerned by the ever-growing AI environmental impact, we hope this work will give rise to future optimizations and new ideas about network design, including the exploration of the aforementioned potential improvements.

\section*{Acknowledgments} 
Part of this work was funded by Hi!PARIS Center on Data Analytics and Artificial Intelligence, by the French National Research Agency (ANR-22-PEFT-0007) as part of France 2030 and the NF-NAI and NF-FITNESS projects and the European Union’s HORIZON Research and Innovation Programme under grant agreement No 101120657, project ENFIELD (European Lighthouse to Manifest Trustworthy and Green AI). 

\nocite{*}
\bibliographystyle{splncs04}
\bibliography{main}

\onecolumn
\appendix

\section{Details on the learning strategies employed}
\label{appendix:hyperparameters}

The training hyperparameters used in the experiments are presented in Table~\ref{tab:training_phases}. %Our code is attached to the supplementary material and will be publically available upon acceptance of the paper.
Our code is available at \url{https://github.com/MaximeGirard/once-for-all-memory-constant}.

\begin{table}[h]
\centering
\setlength{\tabcolsep}{6pt}
\begin{tabular*}{\textwidth}{@{\extracolsep{\fill}}lccccc}
\toprule
\textbf{Task} & \textbf{Phase} & \textbf{Epochs} & \textbf{LR} & \textbf{Warmup Epochs} & \textbf{Warmup LR} \\
\midrule
Teacher & - & 180 & 3.0e-2 & 0 & 3.0e-3 \\
\midrule
Kernel & 1 & 120 & 3.0e-3 & 5 & 3.0e-3 \\
\midrule
Depth & 1 & 25 & 2.5e-4 & - & - \\
\cmidrule{2-6}
 & 2 & 120 & 7.5e-4 & 5 & 7.5e-5 \\
\midrule
Expand & 1 & 25 & 2.5e-4 & - & - \\
\cmidrule{2-6}
 & 2 & 120 & 2.5e-4 & 5 & 7.5e-5 \\
\bottomrule
\\
\end{tabular*}
\caption{Training phases and hyperparameters for each training phase.}
\label{tab:training_phases}
\end{table}
\section{Reference configuration}
\label{appendix:reference_configuration}

We establish a reference configuration with particular parameters for the expand ratio factor, kernel size, and input image size in order to implement this approach.

The reference configuration chosen is :
\begin{itemize}
    \item Depth of 4 for all stages,
    \item Kernel size of 7 for all stages,
    \item Expand ratio of 4 for all stages,
    \item Image resolution of 224.
\end{itemize}

Since these are the values typically found during search and the range around which we want to balance our model, we select values for all parameters that are at their maximum. Naturally, when the chosen constraint is important, a number of parameters must be below the maximum value in order to meet the constraint; however, in most cases, this will result in lower memory usage than when the maximum value is used, which satisfies our requirement.
\section{Graphs for some configurations}
\label{app:graphs}

\begin{figure*}[t]
    \centering
    \begin{subfigure}{0.49\textwidth}
        \centering
        \includegraphics[width=0.5\textwidth]{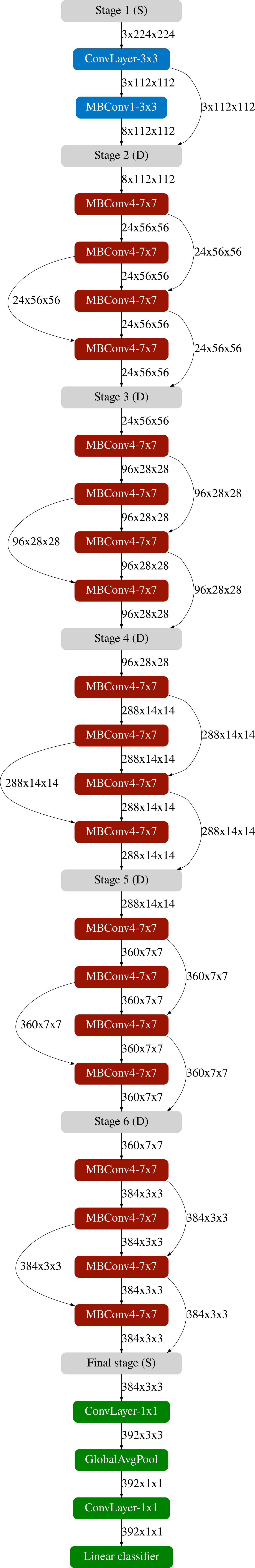}
        \caption{MOOFA - No constraint.}
    \end{subfigure}
    \begin{subfigure}{0.49\textwidth}
        \centering
        \includegraphics[width=0.5\textwidth]{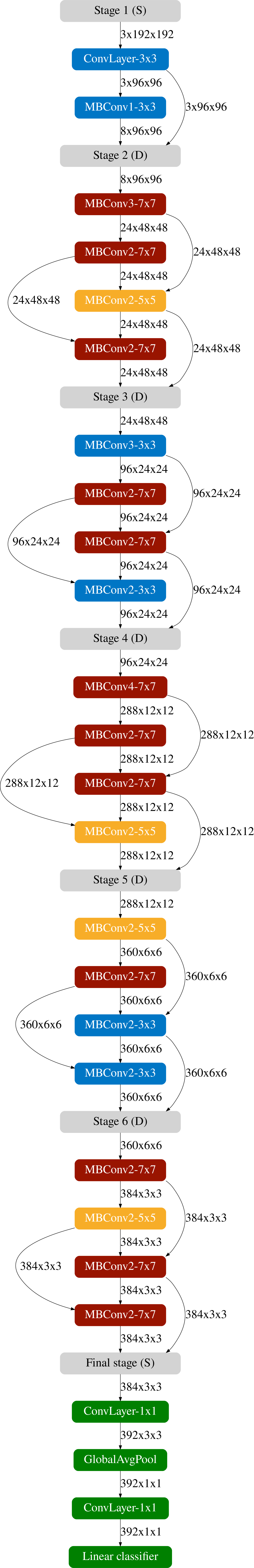}
        \caption{MOOFA - Under 350k constraint.}
    \end{subfigure}
    % \begin{minipage}{0.3\textwidth}
    %     \centering
    %     \vspace{0.24\textwidth}
    %     \includegraphics[width=0.8\linewidth]{figures/MC_600.pdf}
    %     \vspace{0.05\textwidth}
    %     \subcaption{Memory-constant OFA \\ Under 600k \victor{A enlever}constraint.}
    % \end{minipage}
        \caption{Subnets graphs of MOOFA supernet under different constraints.} 
    \label{fig:graphs}
\end{figure*}

Fig.~\ref{fig:graphs} depicts the graphs of subnets for our MOOFA supernet under different constraints. MB Blocks are depicted with format MBConv\{expansion ratio\}-\{kernel size\}. Color refers to kernel size.

% \begin{figure*}[t]
%     \centering
%     \begin{minipage}{0.3\textwidth}
%         \centering
%         \includegraphics[width=0.8\linewidth]{figures/MC_ref.pdf}
%         \vspace{0.05\textwidth}
%         \subcaption{Memory-constant OFA \\ No constraint.}
%     \end{minipage}
%     \begin{minipage}{0.33\textwidth}
%         \centering
%         \vspace{0.18\textwidth}
%         \includegraphics[width=0.7\linewidth]{figures/MC_350.pdf}
%         \vspace{0.0\textwidth}
%         \subcaption{Memory-constant OFA \\ Under 350k constraint.}
%     \end{minipage}
%     % \begin{minipage}{0.3\textwidth}
%     %     \centering
%     %     \vspace{0.24\textwidth}
%     %     \includegraphics[width=0.8\linewidth]{figures/MC_600.pdf}
%     %     \vspace{0.05\textwidth}
%     %     \subcaption{Memory-constant OFA \\ Under 600k \victor{A enlever}constraint.}
%     % \end{minipage}

%         \caption{Subnets graphs for various OFA supernets and constraints. \\ MB Blocks are depicted with format MBConv\{expansion ratio\}-\{kernel size\}. Color refers to kernel size.} 
%     \label{fig:graphs}

% \end{figure*}

\end{document}